%% file: main.tex
\documentclass[10pt,twocolumn,letterpaper]{article}

\usepackage{cvpr}              


\definecolor{cvprblue}{rgb}{0.21,0.49,0.74}
\usepackage[pagebackref,breaklinks,colorlinks,allcolors=cvprblue]{hyperref}

\usepackage{graphicx}       
\usepackage{amsmath}        
\usepackage{subcaption}
\usepackage{adjustbox}
\usepackage{algorithm}
\usepackage{algorithmic}
\usepackage{multirow}
\usepackage{makecell}
\usepackage{hhline}

\def\paperID{4984} 
\def\confName{CVPR}
\def\confYear{2025}

\title{3D-HGS: 3D Half-Gaussian Splatting
\thanks{This work was supported in part  by NSF grant 2038493, ONR grant N00014-21-1-2431, NIH grant R01CA240771 from NCI, and U.S. Department of Homeland Security grant 22STESE00001-03-02. The views and conclusions contained in this document are those of the authors and should not be interpreted as necessarily representing the official policies, either expressed or implied, of the U.S. Department of Homeland Security.}}

\author{
\begin{tabular*}{0.8\textwidth}{@{\extracolsep{\fill}}cccc}
    Haolin Li & Jinyang Liu & Mario Sznaier & Octavia Camps 
  \end{tabular*}\\[1ex]
   \{li.haoli, liu.jiyan, m.sznaier, o.camps\}@northeastern.edu\\
Northeastern University \\
Boston, MA
}

\begin{document}
\maketitle
\input{sec/0abstract}    
\input{sec/1intro}

\input{sec/2relatedwork}

\input{sec/3method}

\input{sec/4experiments}
\input{sec/5conclusion} 
{ 
    \small
    \bibliographystyle{ieeenat_fullname}
    \bibliography{main}
}
\newpage
\input{sec/X_suppl}

\end{document}

%% file: sec/0abstract.tex
\begin{abstract}
Photo-realistic image rendering from 3D scene reconstruction has advanced significantly with neural rendering techniques. Among these, 3D Gaussian Splatting (3D-GS) outperforms Neural Radiance Fields (NeRFs) in quality and speed but struggles with shape and color discontinuities. We propose 3D Half-Gaussian (3D-HGS) kernels as a plug-and-play solution to address these limitations. Our experiments show that 3D-HGS enhances existing 3D-GS methods, achieving state-of-the-art rendering quality without compromising speed.

 More demos and code are available at \url{https://lihaolin88.github.io/CVPR-2025-3DHGS}.
\end{abstract}

%% file: sec/1intro.tex
\section{Introduction}
\label{sec:intro}

\begin{figure}
\begin{subfigure}[b]{0.45\textwidth}
        \centering
        \includegraphics[width=\textwidth]{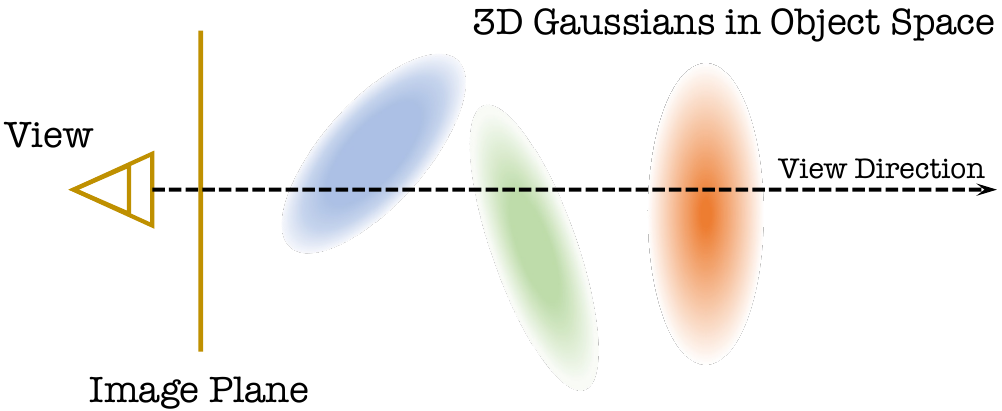}
        \caption{3D Gaussian kernels}
        \label{fig:left}
    \end{subfigure}
    \hfill
    \begin{subfigure}[b]{0.45\textwidth}
        \centering
        \includegraphics[width=\textwidth]{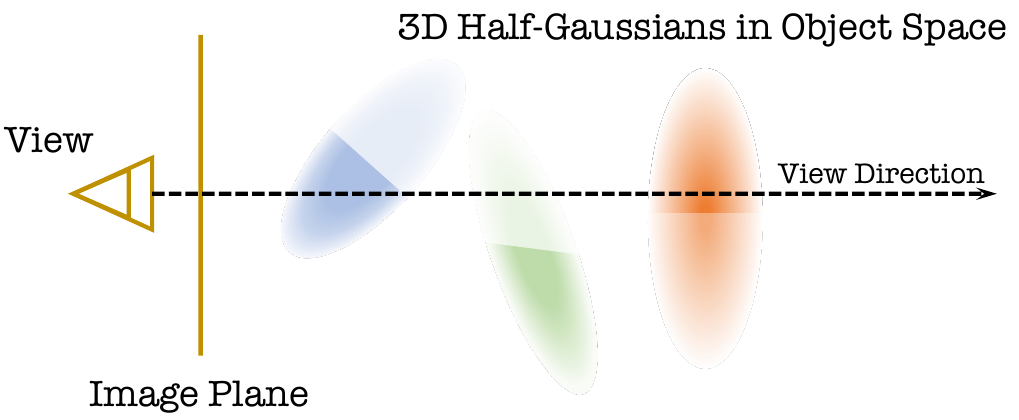}
        \caption{3D Half-Gaussian kernels}
        \label{fig:right}
    \end{subfigure}
    \caption{Illustration of the 3D-GS kernels and the proposed 3D Half-Gaussian kernels, where each half of the kernel is allowed to have different opacity parameters.}
    \vspace{-0.5cm}
    \label{fig:HG}
\end{figure}

The pursuit of photo-realistic and real-time rendering of 3D scenes is a core research focus in both academic and industrial sectors, with wide-ranging applications including virtual reality~\cite{li2023animatable}, media production~\cite{ren2023dreamgaussian4d}, autonomous driving ~\cite{yan2024street, zhao2024tclc}, and extensive scene visualization ~\cite{turki2022mega,lin2024vastgaussian,martin2021nerf}. Traditionally, meshes and point clouds have been the preferred methods for 3D scene representations due to their explicit compatibility with fast GPU/CUDA-based rasterization techniques. However, these methods often result in reconstructions of lower quality and renderings plagued by various artifacts.
In contrast, recent advancements in Neural Radiance Fields (NeRF)~\cite{mildenhall2021nerf}  introduced continuous scene representations leveraging  Multi-Layer Perceptron architectures (MLP). This approach optimizes novel-view synthesis through volumetric ray-marching techniques, providing significantly more realistic renderings. However, NeRF methods are characterized by their slow speed~\cite{gao2022nerf}. 

\begin{figure}
    \centering
    \begin{subfigure}[b]{0.48\textwidth}  
        \centering
        \includegraphics[width=\textwidth]{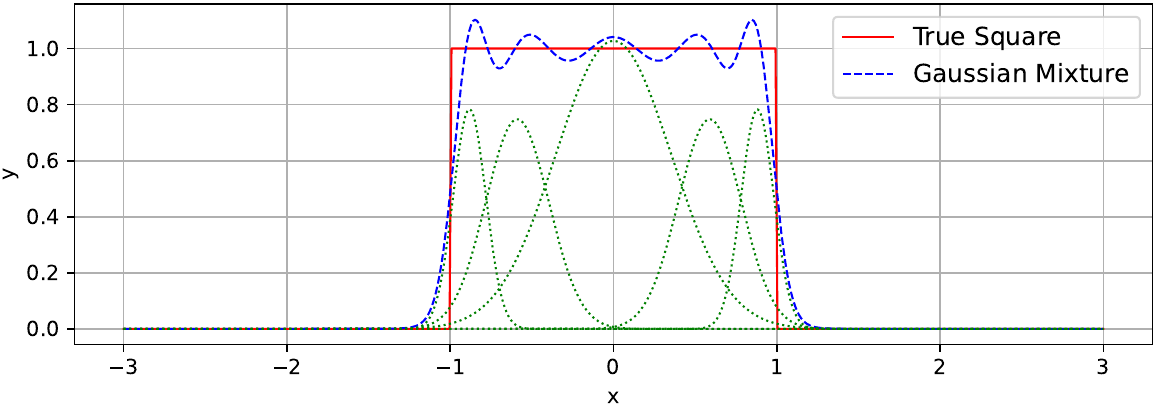}
        \caption{Five Gaussians fitting a square}
        \label{fig:sqleft}
    \end{subfigure}
    \begin{subfigure}[b]{0.48\textwidth} 
        \centering
        \includegraphics[width=\textwidth]{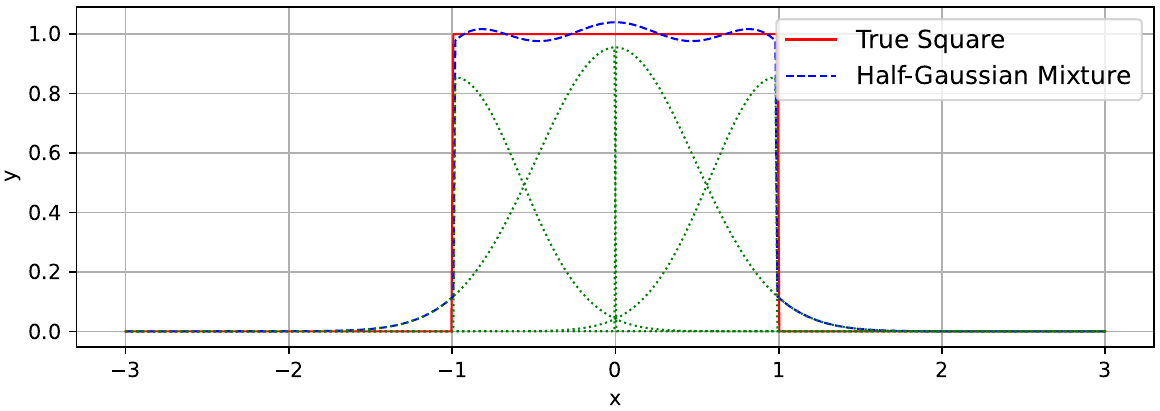}
        \caption{Three HGs fitting a square}
        \label{fig:sqright}
    \end{subfigure}
    \begin{subfigure}[b]{0.48\textwidth}  
        \centering
        \includegraphics[width=\textwidth]{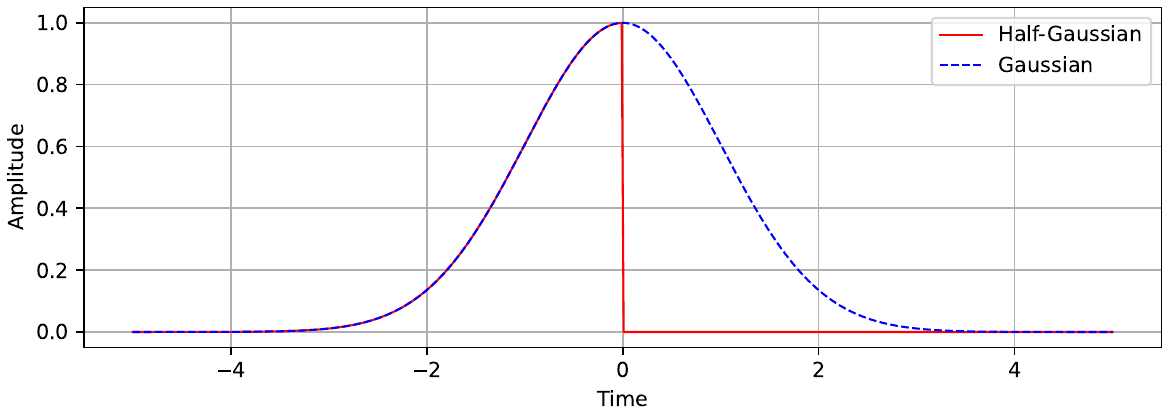}
        \caption{Gaussian and Half-Gaussian in spatial domain}
        \label{fig:sqleft}
    \end{subfigure}
    \begin{subfigure}[b]{0.48\textwidth} 
        \centering
        \includegraphics[width=\textwidth]{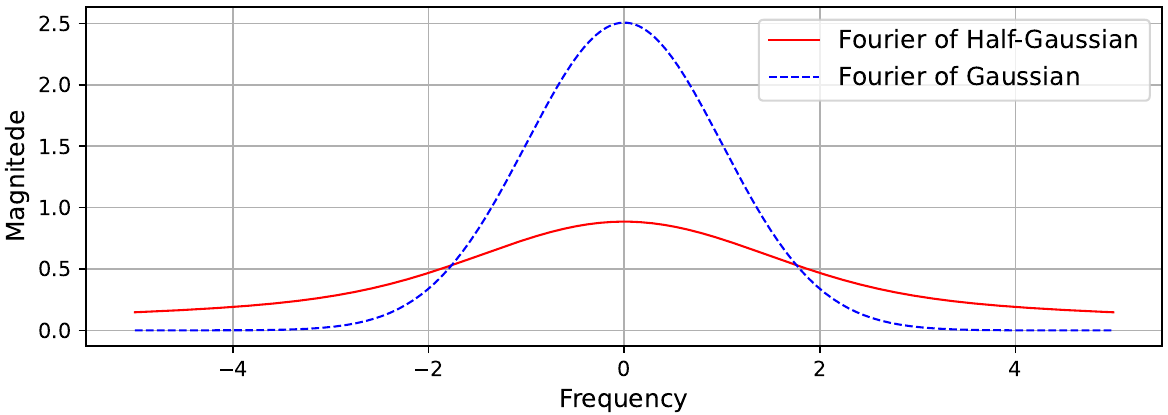}
        \caption{Gaussian and Half-Gaussian in Frequency domain}
        \label{fig:sqright}
    \end{subfigure}
    \caption{\textbf{Comparison of Half-Gaussian and Gaussian Kernels fitting a square function and their Fourier Transforms. }(a): fitting a square function with 5 Gaussian kernels, and (b): fitting a square with 4 Half-Gaussian kernels. When approximating sharp edges, the Half-Gaussian kernels achieve a lower error loss (1.85) compared to Gaussian kernels (2.97). Figures (c) and (d) illustrate the Gaussian and Half-Gaussian kernels in both the spatial and frequency domains, where the Half-Gaussian demonstrates a higher bandwidth than the Gaussian kernel, indicating its superior ability to capture high-frequency components. }
    \label{fig:square}
\end{figure}

Recently, 3D Gaussian splatting ({3D-GS})~\cite{kerbl20233d} has emerged as a state-of-the-art approach, outperforming existing methods in terms of both rendering quality and speed. The concept of using 3D Gaussians to parameterize a scene dates back to the early 2000s~\cite{zwicker2002ewa,zwicker2001surface}.  This technique models a 3D scene with a collection of 3D Gaussian reconstruction kernels parameterized by 59 parameters representing location, scale, orientation, color, and opacity of the kernel. Initially, each Gaussian is derived from a point in a Structure from Motion (SfM) reconstruction. These parameters are subsequently refined through a dual process involving the minimization of image rendering loss and adaptive kernel density adjustment. The efficiency of 3D-GS is enhanced by a GPU-optimized, tile-based rasterization method, enabling real-time rendering of complex 3D scenes.
Employing 3D Gaussians as reconstruction kernels simplifies the volumetric rendering process and facilitates the integration of a low-pass filter within the kernel. This addition effectively mitigates aliasing during the resampling of scene data to screen space~\cite{zwicker2001ewa,zwicker2002ewa}.

\begin{figure}
    \centering
    \includegraphics[width=0.48\textwidth]{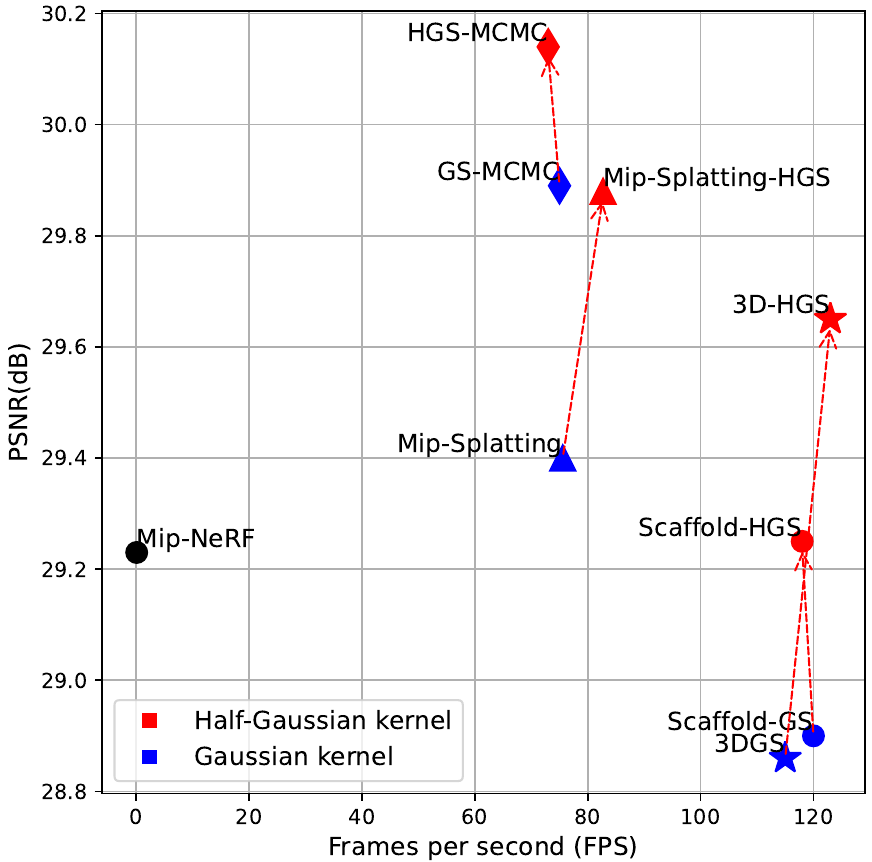}
    \caption{{Performance (PSNR$\uparrow$) versus rendering speed for several state-of-the-art methods \cite{kerbl20233d,lu2023scaffold,kheradmand20243d,barron2021mip,yu2024mip} with Gaussian kernels and the proposed half-Gaussian kernels on the Mip-NeRF360 dataset~\cite{barron2021mip}. In all cases, using half-Gaussian kernels resulted in significant PSNR improvements, with similar or better rendering speed than the corresponding 3D Gaussian-based method. }}
    \label{fig:comparisonHG}
\end{figure}

While 3D-GS benefits from employing  3D Gaussians, using a Gaussian kernel can be inefficient and lead to inaccuracies when modeling discontinuous functions, which are common in 3D scenes at  object boundaries and texture-rich areas. As Fig.~\ref{fig:square}(a) shows, using Gaussian kernels leads to a Gibbs phenomenon  with high peaks overshooting and undershooting the function value and non-vertical transitions (i.e. blurry edges). 
To address this problem, we propose the use of 3D-Half-Gaussian Splatting (3D-HGS), which employs a 3D-Half Gaussian as a novel reconstruction kernel. As seen in Fig.~\ref{fig:square}(b), using  this kernel  significantly reduces the Gibbs oscillatory effect and better fits the vertical discontinuities. This can be explained by the fact that Half-Gaussians have a higher content at higher frequencies  than full Gaussians do  (Fig.~\ref{fig:square} (c) and (d)). 

A pair of  3D half-Gaussians can be  easily represented by adding the parameters of a vector normal to a  plane splitting a 3D Gaussian through its center, and allowing each half to have different opacity values (Fig.~\ref{fig:HG}(b)). By introducing  this plane, the kernel can efficiently capture high-frequency information at discontinuities, while preserving the essential characteristics of the original 3D Gaussian kernel (Fig.~\ref{fig:square}). This preservation is facilitated by the symmetric pairing of 3D Half-Gaussians, which also allows to  seamlessly represent  full 3D Gaussians (see for example the center Gaussian in Fig.~\ref{fig:square}(b)). 

Thus, the proposed 3D-Half-Gaussian kernel preserves the key parameters in the original {3D-GS} and provides the capability to be easily applied to existing 3D Gaussian kernel-based methods as a plug-and-play kernel.  
Our experiments (see Fig.~\ref{fig:comparisonHG} and section~\ref{sec:exp}) show that using the proposed 3D Half-Gaussian as the reconstruction kernel achieves state-of-the-art (SOTA) rendering quality performance on MipNeRF-360, Tanks\&Temples, and Deep Blending datasets, with similar or better rendering speed than the corresponding 3D Gaussian-based methods.  

Our main contributions are: (1) We introduce a novel plug-and-play reconstruction kernel, the 3D Half-Gaussian, designed to enhance the performance of 3D Gaussian Splatting (3D-GS). (2) Our proposed kernel achieves state-of-the-art novel view synthesis performance across multiple datasets without compromising rendering frame rate. (3) We demonstrate the versatility and effectiveness of our kernel by applying it to other state-of-the-art methods, showcasing its broad applicability. Our code will be available on our GitHub project page.

%% file: sec/2relatedwork.tex
\section{Related Work}
\label{sec:relatedwork}

3D reconstruction and Novel View Synthesis (NVS) have long been key goals in computer vision. NeRFs \cite{mildenhall2021nerf} have significantly  advanced NVS, enabling highly realistic image synthesis from new viewpoints. More recently, 3D-GS has set new state-of-the-art benchmarks in this field. This section reviews both historical and recent developments in 3D reconstruction and NVS, with a detailed analysis of 3D-GS, its methods, achievements, and impact on the field.

\subsection{Novel View Synthesis}
Before the advent of NeRFs, Multi-View Stereo (MVS)~\cite{seitz2006comparison} and Structure from Motion (SfM) were commonly utilized for reconstructing 3D scenes from multiple viewpoint images. MVS relies on feature extraction from diverse images to correlate viewpoints and produce a final reconstruction, typically represented as a colored mesh grid or point cloud. However, due to its reliance solely on image features, achieving high-quality reconstructions can be challenging.
SfM~\cite{schonberger2016structure}, employs multi-view images to generate a point cloud representing the 3D scene. In contrast to MVS, SfM excels in accurately estimating camera poses for different images, while MVS provides more precise 3D estimations of the scene. Notably, SfM's simultaneous estimation of camera positions and point clouds makes it a preferred pre-processing step in recent advanced NVS methods.

NeRFs \cite{mildenhall2021nerf}, stand as a significant milestone in NVS, demonstrating remarkable achievements in tasks such as image and view synthesis alongside scene reconstruction. Unlike previous methods, a Nerf represents the 3D scene using a radiance field, treating the entire scene as a continuous function parameterized by position.

3D-GS \cite{kerbl20233d}, is a recently introduced NVS technique. This method boasts a remarkable reduction in both training and rendering times, surpassing the Nerf methods.
Diverging from its predecessors, 3D-GS does not rely on training neural networks or any type of network architecture. Instead, it initiates from a point cloud. Unlike treating each point discretely, 3D-GS conceptualizes them as 3D Gaussian entities, each possessing a unique size and orientation,  and  spherical harmonics to depict their color. Subsequently, during the splatting stage, these 3D Gaussians are projected onto a 2D plane, with their appearances accumulated to generate renderings for a given viewing angle.

\subsection{Splatting methods}

The original concept of splatting was introduced by Westover~\cite{westover1989interactive,westover1990footprint}, and improved by Zwicker et. al~\cite{zwicker2001ewa,zwicker2002ewa,zwicker2001surface}. Recently, the 3D-GS technique has achieved great success in photo-realistic neural rendering~\cite{kerbl20233d,wu2024recent}. 

Although 3D-GS has attained state-of-the-art performance in terms of rendering speed and quality, opportunities for further enhancements remain. Various studies~\cite{chen2024survey, dalal2024gaussian} have introduced modifications to the original framework. For instance, Mip-splatting~\cite{yu2024mip} limits the frequency of the 3D representation to no more than half the maximum sampling frequency, as determined by the training images. Analytic-Splatting~\cite{liang2024analytic} employs a logistic function to approximate the Gaussian cumulative distribution function, thus refining each pixel’s intensity response to minimize aliasing. Similarly, SA-GS~\cite{song2024sa} adapts  during testing a 2D low-pass filter based on rendering resolution and camera distance. Scaffold-GS~\cite{lu2023scaffold} employs voxel grids for initializing 3D scenes and utilizes Multi-Layer Perceptrons (MLPs) to constrain and learn voxel features. Following a similar paradigm, SAGS~\cite{ververas2024sags} incorporates a graph neural network to capture structural relationships between individual Gaussians, preserving geometric consistency across neighboring regions during rendering. FreGS~\cite{zhang2024fregs} introduces frequency-domain regularization to the rendered 2D images to enhance the recovery of high-frequency details. 2D-GS~\cite{huang20242d} aligns 3D scenes with 2D Gaussian kernels to improve surface normal representations. MCMC-GS~\cite{kheradmand20243d} applied Stochastic Gradient Langevin Dynamics (SGLD) to iteratively refine Gaussian positions. Lastly, GES~\cite{hamdi2024ges} employs generalized exponential kernels to reduce the memory required to store 3D information.

The two papers most closely related to ours are 2D-GS~\cite{huang20242d} and GES~\cite{hamdi2024ges}, as they also focus on switching the reconstruction kernels. However, GES did not change the rasterizer but learned a scaling factor for each of the points to approximate different kernels. This method faces difficulties in complex 3D scenes. On the other hand, instead of doing a volumetric rendering as in the original 3D-GS, 2D-GS tries to do a surface rendering. Our method still learns a volumetric rendering of scenes to retain the rendering performance while enabling accurate modeling of discontinuous functions.

%% file: sec/3method.tex
\section{Method}

\begin{figure*}[h]
    \centering
    \begin{subfigure}[b]{0.80\textwidth}
        \centering
        \includegraphics[width=\textwidth]{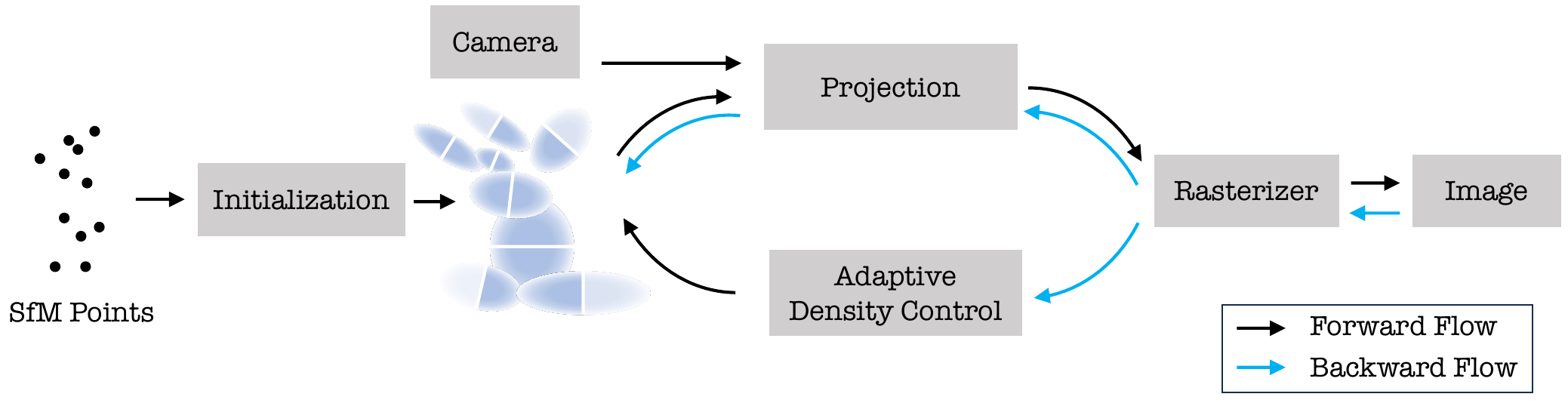}
        \caption{The training and rendering pipeline for 3D Half-Gaussian}
        \label{fig:left}
    \end{subfigure}
    
    \begin{subfigure}[b]{0.22\textwidth}  
        \centering
        \includegraphics[width=\textwidth]{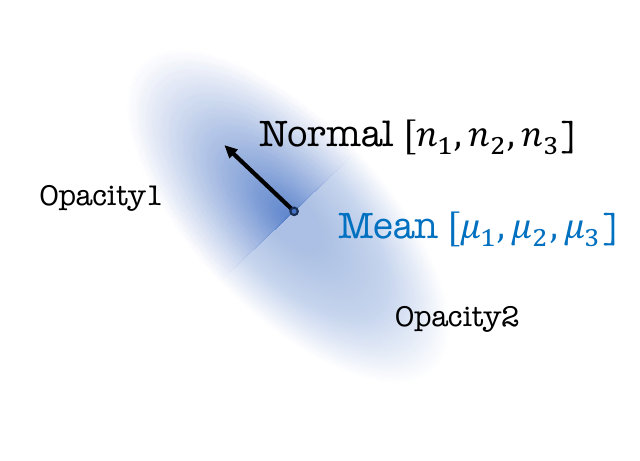}
        \caption{3D Half-Gaussian pair}
        \label{fig:HGleft}
    \end{subfigure}
    \begin{subfigure}[b]{0.56\textwidth} 
        \centering
        \includegraphics[width=\textwidth]{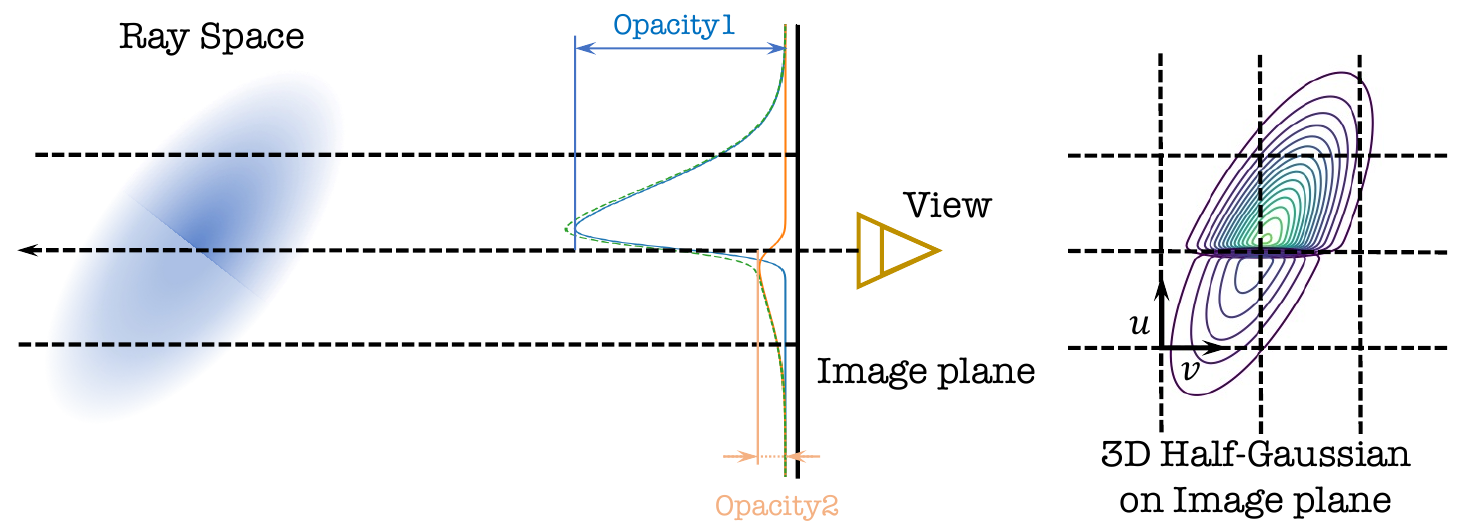}
        \caption{Mapping the Half-Gaussian to image plane in the ray space}
        \label{fig:HGright}
    \end{subfigure}
    \caption{Illustration of the 3D-HG kernel, and the mapping of a pair of 3D Half-Gaussians to a 2D image.}
    \label{fig:HG}
\end{figure*}

3D-GS \cite{kerbl20233d} models  3D scenes using 3D Gaussian kernels \cite{zwicker2001surface,zwicker2002ewa}. However, these kernels frequently encounter difficulties in accurately modeling 2D/3D discontinuous functions, which are prevalent in 3D scenes at edges, corners of objects, and texture-rich areas. The inefficiencies of the 3D Gaussian kernel in representing shape and color discontinuities can compromise the model's effectiveness. To overcome these limitations, we propose the use of a Half-Gaussian kernel for novel view synthesis as illustrated in Fig.~\ref{fig:HG}. Section \ref{Preliminaries} presents an overview of the foundational concepts relevant to 3D Gaussian Splatting. In Section \ref{3dhg}, we introduce the 3D half-Gaussian kernel, followed by an in-depth description of the 3D-HGS rasterizer in Section \ref{3.3}. Finally, Section \ref{3.4} provides a detailed account of the splatting process.

\subsection{Preliminaries}\label{Preliminaries}

3D-GS \cite{kerbl20233d} represents the 3D scenes  with parameterized antisotropic 3D Gaussian kernels \cite{zwicker2001surface,zwicker2002ewa}. It starts from an initial set of 3D Gaussians located at a set of sparse  points  obtained using a Structure-from-Motion (SfM) step. Then, 3D Gaussians are mapped to 2D images through a GPU-specified tile-based rasterization. The parameters of these Gaussians are optimized, pruned, and added based on a loss function on the rendered images and ground-truth images.

Each of the 3D Gaussians is parameterized by their position (mean) $\mu$, covariance-related scaling matrix, opacity $\alpha$, and Spherical harmonics coefficients for color $c$. 
A 3D elliptical Gaussian $G_{\mathbf{\Sigma}}(\textbf{x})$ centered at a point $\mathbf{\mu}$ with  covariance matrix  $\mathbf{\Sigma}$ is given by:
\begin{equation}
    G_{\mathbf{\Sigma}}(\textbf{x}-\mathbf{\mu}) = \frac{1}{(2 \pi)^{3/2}|\mathbf{\Sigma}|^{\frac{1}{2}}}e^{-\frac{1}{2}(\textbf{x}-\mathbf{\mu})^T\mathbf{\Sigma}^{-1}(\textbf{x}-\mathbf{\mu})}
\end{equation} 
where $\textbf{x}$ and $\mathbf{\mu}$ are the column vectors $[x,y,z]^T$ and $[\mu_x,\mu_y,\mu_z]^T$, respectively, and $\mathbf{\Sigma}$ is a positive definite $3\times3$ matrix. 
The covariance under the world coordinate system $\mathbf{\Sigma}^w$ is further parameterized by a scaling matrix $S$ and a rotation matrix $R$ to maintain its positive definite property:
\begin{equation}
\boldsymbol{\Sigma^w} = \boldsymbol{R}\boldsymbol{S}\boldsymbol{S}^T\boldsymbol{R}^T\label{}
\end{equation}

Given a viewing transformation $W$ and the Jacobian of the affine approximation of the projective
transformation $J$, the covariance matrix $\boldsymbol{\Sigma}$ in the camera coordinate system is given by:

\begin{equation}
\boldsymbol{\Sigma} = JW\boldsymbol{\Sigma}^wW^TJ^T\label{eq:cov}
\end{equation}
 
As in the rendering function we will learn the opacity parameter $\alpha$, we can merge the constant term $\frac{1}{(2 \pi)^{3/2}|\mathbf{\Sigma}|^{\frac{1}{2}}}$ into $\alpha$.  
Integrating a normalized 3D Gaussian along one coordinate axis results in a normalized 2D Gaussian. Thus, 3D Gaussians $G(\mathbf{x})$ can be efficiently transformed to 2D Gaussians $\hat{G}(\hat{\mathbf{x}})$ on the image plane using a ray coordinate system representation \cite{zwicker2001ewa}:
\begin{equation}\label{eq:5}
    \int_{\mathbb{R}}G_{\mathbf{\Sigma}}(\mathbf{x}-\mathbf{\mu})dz = \hat{G}_{\mathbf{\hat{\Sigma}}}(\hat{\mathbf{x}}-\mathbf{\hat{\mu}})  
\end{equation}
where $\hat{\mathbf{x}}=[x,y]^T$, $\hat{\mathbf{\mu}}=[\mu_x,\mu_y]^T$, and  the covariance ${\mathbf{\hat{\Sigma}}}$ can be easily obtained by taking the top-left $2\times2$ sub-matrix of the transformed $\mathbf{\Sigma}$: 

\begin{equation}\label{eq:6}
  \mathbf{\Sigma} =   \begin{pmatrix}
a & b & c\\
b & d & e \\
c & e & f \\
\end{pmatrix}  \Leftrightarrow 
\begin{pmatrix}
a & b \\
b & d  \\
\end{pmatrix} = {\mathbf{\hat{\Sigma}}}
\end{equation}
Finally, the pixel values on the 2D image are obtained by $\alpha$-blending:
\begin{equation} 
C=\sum_{i \in \mathcal{N}}c_i\sigma_i \prod_{j=1}^{i-1}(1-\sigma_j), \ \ \sigma_i=\alpha_i \hat{G}(\boldsymbol{\hat{x} - \hat{\mu}})
\end{equation}
where $\hat{\mathbf{x}}$ is the queried pixel position and  $\mathcal{N}$ is the set of sorted 2D Gaussians associated with $\hat{\mathbf{x}}$.

\subsection{3D Half-Gaussian Kernel}\label{3dhg}

In this section, we provide a detailed description of the proposed kernel. We begin by giving the definition of the half Gaussian kernel and how to parameterize it.

We propose to use  3D Half-Gauusians kernels, where each half can have different opacity values $\alpha_1$ and $\alpha_2$. The half Gaussians are obtained by splitting a 3D Gaussian  with a plane through its center.  Note that this representation includes as a special case the 3D-GS representation in the case when $\alpha_1=\alpha_2$. However,  incorporating a planar surface into the kernel allows to represent sharp changes, such as edges and textures, as well as planar surfaces, more accurately. As seen below, the increase in the number of learned parameters is very modest: the 3D normal of the splitting plane and an additional opacity term.  {Furthermore, it should be noted that the plane normal can be stored using the normal field, which is available but not used, in the original 3DGS kernel implementation. Thus, 3DHGS effectively increases the memory requirements only by an extra opacity coefficient, while it increases computational cost by a multiplication by a scaling factor.}

Formally, a 3D Half-Gaussian kernel is described by: 
\begin{equation}
    HG_\mathbf{\Sigma}(\textbf{x}-\mu) = 
    \left\{
    \begin{array}{ll}
    e^{-\frac{1}{2}(\textbf{x}-\boldsymbol{\mu})^T\boldsymbol{\Sigma}^{-1}(\textbf{x}-\boldsymbol{\mu})} &  \mathbf{n}^T(\mathbf{x} - \mu) \ge 0\\
    0 & \mathbf{n}^T(\mathbf{x} - \mu) <0\\
    \end{array}
    \right. 
\end{equation}
where $\mathbf{n} = [n_1,n_2,n_3]^T$ is the normal of the splitting plane through the full Gaussian center, represented as a column vector, while the complementary half Gaussian is obtained by simply changing the sign of the normal. The integration of a 3D Half Gaussian along the $z$ axis yields a similar result as Eq.~\ref{eq:5}, except for a scaling factor related to the normal of the splitting plane:
\begin{equation}\label{eq:7}
    \int_{\mathbf{n}^T(\mathbf{x}-\mu) \ge 0}{HG}_{\mathbf{\Sigma}}(\mathbf{x} - \mu)dz =
    \frac{1}{2}I(x,y)\hat{G}_{\mathbf{\hat{\Sigma}}}(\hat{\mathbf{x}}-\mathbf{\hat{\mu}})
\end{equation}
where
\vspace{-0.3cm}
\begin{equation}\label{eq:8}
I(x,y) = \text{erfc}\left( -\frac{
(n_1x+n_2y)/n_3 + \mu_{z|xy}}{\sqrt{2}\sigma_{z|xy}}
 \right)
\end{equation}
and $\mu_{z|xy}$, and $\sigma_{z|xy}$ refer to the mean and standard deviation for the distribution of $z$,  conditioned on given $x, y$, i.e. $P(z|x,y) \sim \mathcal{N}(\mu_{z|xy},\sigma_{z|xy})$, respectively. 

Eq.~\ref{eq:7} and Eq.~\ref{eq:8} give a closed-form solution to calculate the integral of a 3D Half-Gaussian. The right-hand side of Eq.~\ref{eq:7} consists of two factors: $I(x,y)$, which is a scaling factor related to the learned normal of the splitting plane, while $\hat{\mathbf{x}}-\mathbf{\hat{\mu}}$ remains the same as in the original 3D Gaussian in Eq.~\ref{eq:5}. For a detailed derivation please refer to the supplementary material.

\subsection {3D Half-Gaussian Rasterization}\label{3.3}

For rasterization, we follow a similar process as in {3D-GS}, adapted to the proposed 3D Half-Gaussian reconstruction kernel. For enhanced parameter efficiency, we jointly represent pairs of Half-Gaussians, since the parameters for mean, rotation, scaling, and color are shared between the two halves, while we learn the orientation of the splitting plane and two opacity parameters, one for each side of the splitting plane.   Thus, the volumetric alpha blending for each pixel on the image can be expressed as :
\begin{equation} 
    C = \sum_{i\in \mathcal{N}}c_i H\hat{G}_i(\hat{\mathbf{x}}- \hat{\mu}) \prod_{j=1}^{i-1}\left(1- H\hat{G}_j (\hat{\mathbf{x}} - \hat{\mu})\right)
\end{equation}
where $\mathcal{N}$ is the sorted 3D Half-Gaussian set for the pixel and $H\hat{G}(x)$ is the integration of both halves of a 3D Half-Gaussian  pair:
\begin{equation}\label{eq:10}
    H\hat{G}(\mathbf{\hat x}- \hat{\mu}) =\frac{1}{2} \left \{2\alpha_2+\left(\alpha_1 -\alpha_2\right)I(x,y)\right\}\hat{G}_{\mathbf{\hat{\Sigma}}}(\hat{\mathbf{x}}-\mathbf{\hat{\mu}}) 
\end{equation}

\subsection{3D Half-Gaussian Splatting}\label{3.4}

\begin{figure}[t]
        \centering
        \includegraphics[width=0.48\textwidth]{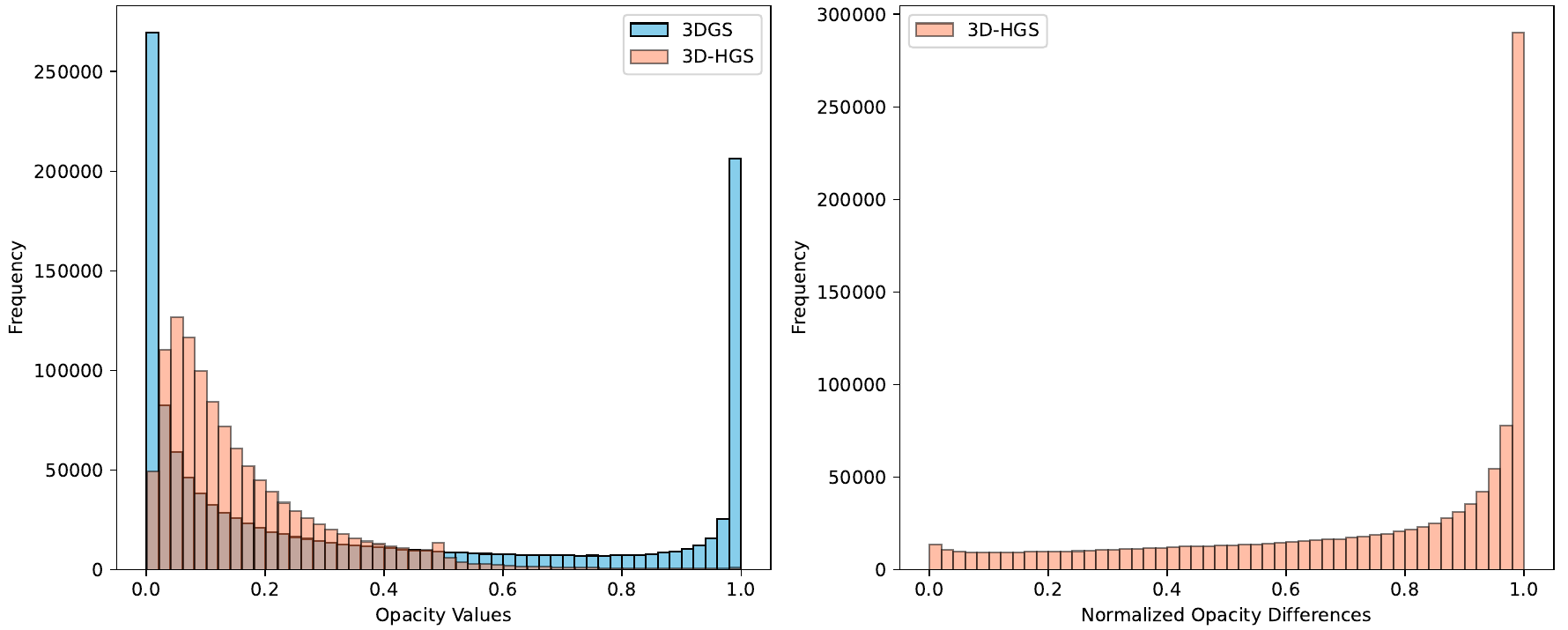}
    \caption{(Left) Histograms of the 3DGS opacity values  and of the 3D-HGS mean opacity values of corresponding halves,  trained on the Bonsai scene. The 3D-HGS mean opacity values   cluster in a lower range than those in the 3DGS, implying that treating both halves identically while rendering increases the number of kernels involved in each tile, slowing down the overall process. {(Right)} Histogram of the  differences between opacity values  within a half Gaussian, normalized by the maximum opacity value within each kernel. Over 75\%  of the 3D-HGS kernels have normalized opacity differences over 0.5, highlighting that each half Gaussian  often represents a distinct effective area in the rendering space. }
   \vspace{-0.5cm}
    \label{fig:hist}
\end{figure}

The 3D half-Gaussian kernel has a splitting plane along with distinct opacity values for each half (Fig.~\ref{fig:HGleft}).  Unlike conventional splatting methods -- which compute the projection shape of a Gaussian as a whole -- the half-Gaussian kernel requires specialized handling. 

Naively  applying standard splatting uniformly across both halves is inefficient  since, in general, a large number of kernels have one of their halves fully transparent. This is supported by Fig.~\ref{fig:hist}, where the left plot shows that the average opacity values per half-Gaussian kernel cluster in a lower range than when using the full Gaussian kernel and the right plot shows that over 75\% of the kernels have normalized opacity differences greater than 0.5.  
Therefore,  treating both halves identically while rendering would increase the number of kernels involved in each tile, slowing down the overall process. 

\begin{figure}[t]
        \centering
        \includegraphics[width=0.48\textwidth]{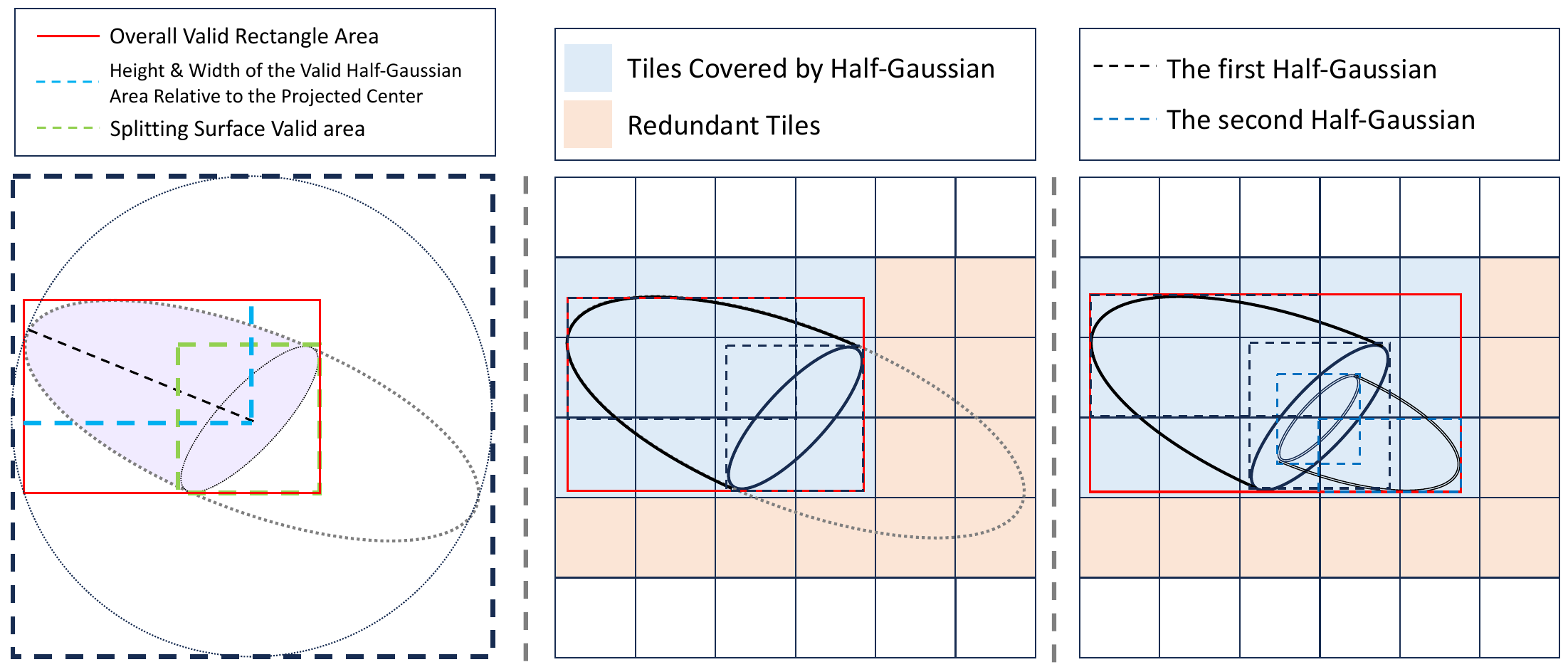}
    \caption{{Left: Method for calculating the bounding rectangle of a half-Gaussian. Middle: Visualization when one half of the Gaussian is transparent. Right: Visualization when each side of the Gaussian has distinct opacity levels.}}
\vspace{-0.8cm}
    \label{fig:HGrendering}
\end{figure}

Thus,  we propose an efficient  3D half-Gaussian splatting technique that independently calculates the valid region for each half as illustrated in Fig.~\ref{fig:HGrendering}. 
Initially, we project the ellipsoidal splitting surface onto the 2D image plane, defining the inner {rectangle} (relative to the projected center) of the projected half-Gaussian {(green dashed rectangle in Fig.~\ref{fig:HGrendering})}. To determine the outer bound, inspired by~\cite{feng2024flashgs}, we compute the height and width of the tangent edges relative to the ellipse’s center {(blue dashed lines in Fig.~\ref{fig:HGrendering})}, thereby establishing the limits of the valid region. This region is encapsulated by the minimal bounding rectangle that fully encloses the projected splitting ellipse along with its tangent edges (red rectangle in Fig.~\ref{fig:HGrendering}).

In practice, if one side of the half-Gaussian is entirely transparent, only the opaque half is splatted, as depicted in Fig.~\ref{fig:HGrendering}, center. When both halves contribute, the valid region is determined by the outer bounding tangent edges and the projected splitting ellipse Fig.~\ref{fig:HGrendering}, right.

%% file: sec/4experiments.tex
\section{Experiments}

\subsection{Experimental Setup}

\textbf{Datasets and Metrics.} 
Following the published literature, we tested our design on  11 (indoor and outdoor) scenes from various datasets: 7 scenes from Mip-nerf360 dataset\cite{barron2021mip}, 2 scenes from Tanks \&Temples~\cite{knapitsch2017tanks}, and 2 scenes from DeepBlending~\cite{hedman2018deep}. 

Consistent with prior studies, we use PSNR, SSIM~\cite{wang2004image} and LPIPS~\cite{zhang2018unreasonable} to measure the performance on each dataset. We provide the averaged metrics over all scenes for each dataset in the main paper and give the full quantitative results for each scene in the Appendix. We also report rendering times and model size. 

\noindent \textbf{Baselines.}
To evaluate the general improvement brought by using a Half-Gaussian kernel in neural rendering, we ran experiments based on four baseline methods: vanilla-{3D-GS}~\cite{kerbl20233d} and three of its extensions {Scaffold-GS~\cite{lu2023scaffold}, Mip-Splatting~\cite{yu2024mip}, and GS-MCMC~\cite{kheradmand20243d}}, where we replaced the reconstruction kernel with the Half-Gaussian kernel. We denote our models as {3D-HGS}, {Scaffold-HGS}, {Mip-Splatting-HGS}, and {HGS-MCMC}, respectively.
We compared performance against  state-of-the-art 3D reconstruction methods: 3D-GS~\cite{kerbl20233d}, Mip-NeRF~\cite{barron2021mip}, 2D-GS~\cite{huang20242d}, Fre-GS~\cite{zhang2024fregs}, Scaffold-GS~\cite{lu2023scaffold}, GES~\cite{hamdi2024ges}, Mip-Splatting~\cite{yu2024mip}, and GS-MCMC~\cite{kheradmand20243d}.

\noindent{\bf Reconstruction Kernels.} We compared the rendering performance of the half-Gaussian kernel against three other reconstruction kernels: the original 3D-GS kernel,  
the 2D Gaussian kernel (2D-GS) and the generalized exponential kernels (GES), which were proposed in \cite{huang20242d} and \cite{hamdi2024ges}, respectively. For a fair comparison, we used the same loss function and training iterations as the original 3D-GS. For details on the implementation of different kernels, please refer to the supplementary material.

\begin{table*}[t]
\caption{\textbf{Quantitative comparison to the SOTA methods on real-world datasets. }}
\begin{adjustbox}{width=\linewidth}
    \begin{tabular}{ l | l l l | l l l | l l l}
 
    Dataset & \multicolumn{3}{c|}{Mip-NeRF360} & \multicolumn{3}{c|}{Tanks\&Temples} & \multicolumn{3}{c}{Deep Blending}  \\
    
    Method& PSNR $\uparrow$ & SSIM$\uparrow$ & LPIPS$\downarrow$ & PSNR$\uparrow$ & SSIM$\uparrow$ & LPIPS$\downarrow$ & PSNR$\uparrow$ & SSIM$\uparrow$ & LPIPS$\downarrow$   \\
    \cline{1-10}
    \cline{1-10}
    \cline{1-10}
    \cline{1-10}
    \cline{1-10}
    \cline{1-10}
    
    Mip-NeRF~\cite{barron2021mip}& 29.23&0.844 &0.207 &22.22 &0.759 &0.257 &29.40 & 0.901&0.245  \\

    2D-GS\cite{huang20242d}& 28.98 &0.867 &0.185 & 23.43&0.845 &0.181 & 29.70 &0.902&0.250 \\
    
    Fre-GS~\cite{zhang2024fregs}& 27.85 &0.826 & 0.209 &23.96 &0.841 &0.183  & 29.93 &0.904 & 0.240 \\

    GES~\cite{hamdi2024ges}&28.69 &0.857& 0.206 &23.35&0.836&0.198 &29.68&0.901&0.252 \\

    \cline{1-10}
    \cline{1-10}
    \cline{1-10}
    \cline{1-10}
    \cline{1-10}
    \cline{1-10}
    3D-GS \cite{kerbl20233d} &28.88 &0.870& 0.182 &23.60 & 0.847&0.181   &29.41&0.903 & 0.243  \\
    \small 3D-\textbf{HGS} (\textbf{Ours}) & 29.66 \textbf{\scriptsize{+0.78}} & 0.873& 0.178  &\cellcolor{yellow!30}{24.45} \textbf{\scriptsize{+0.85}} & 0.857 & 0.169 & 29.76 \textbf{\scriptsize{+0.35}}& \cellcolor{yellow!30}0.905&\cellcolor{yellow!30}0.242 \\
    
    \cline{1-10}
    \cline{1-10}
    \cline{1-10}
    \cline{1-10}
    \cline{1-10}
    \cline{1-10}
    Scaffold-GS~\cite{lu2023scaffold} &28.95& 0.848&0.220  &23.96 &0.853 & 0.177 &\cellcolor{orange!30}30.21 &\cellcolor{orange!30}0.906 & 0.254 \\
    \small Scaffold-\textbf{HGS}(\textbf{Ours}) & 29.25 \textbf{\scriptsize{+0.30}} & 0.867 & 0.186 &24.42 \textbf{\scriptsize{+0.46}}& \cellcolor{yellow!30}0.859 & 0.162  & \cellcolor{red!30}30.36 \textbf{\scriptsize{+0.15}} &  \cellcolor{red!30}0.910& \cellcolor{red!30}0.240  \\

    \cline{1-10}
    \cline{1-10}
    \cline{1-10}
    \cline{1-10}
    \cline{1-10}
    \cline{1-10}
    Mip-Splatting~\cite{yu2024mip} &29.39&0.880&\cellcolor{yellow!30}0.162&23.75&0.857&\cellcolor{yellow!30}0.157 &29.46&0.903&0.243\\
    \small Mip-Splatting-\textbf{HGS}(\textbf{Ours}) &\cellcolor{yellow!30}29.88 \textbf{\scriptsize{+0.49}} &\cellcolor{yellow!30}0.881&\cellcolor{orange!30}0.160& \cellcolor{orange!30}24.53 \textbf{\scriptsize{+0.78}} &\cellcolor{red!30}0.865&\cellcolor{orange!30}0.145& 29.61 \textbf{\scriptsize{+0.15}} &0.901&\cellcolor{orange!30}0.241\\

    \cline{1-10}
    \cline{1-10}
    \cline{1-10}
    \cline{1-10}
    \cline{1-10}
    \cline{1-10}
    
    GS-MCMC~\cite{kheradmand20243d} &\cellcolor{orange!30}29.89 &\cellcolor{red!30}0.900 &0.190 &24.29 &\cellcolor{orange!30}0.860 &0.190 &29.67 &0.890 &0.320 \\
    \small \textbf{HGS}-MCMC(\textbf{Ours}) &\cellcolor{red!30}30.13 \textbf{\scriptsize{+0.24}}  & \cellcolor{orange!30}0.886 &\cellcolor{red!30}0.158   & \cellcolor{red!30}25.08 \textbf{\scriptsize{+0.77}}  & 0.841 & \cellcolor{red!30}0.144 &  \cellcolor{yellow!30}29.80 \textbf{\scriptsize{+0.13}} & 0.898&0.245  \\

    \end{tabular}
\end{adjustbox}
\label{Tab:QuantaResults}
\end{table*}

\noindent \textbf{Implementation.} For our implementation, we utilize the three (unused) normal parameters in 3D-GS to represent the normal vector of the splitting plane. Additionally, we learn one opacity for each half of the Gaussian. This results in memory increasing by only one additional parameter for each reconstruction kernel compared to the 3D-GS. The forward and backward passes of the rasterizer are modified based on the vanilla 3D-GS and Eq.~\ref{eq:10}. For 3D-HGS and Mip-Splatting-HGS, we adhered to the training settings and hyperparameters used in \cite{kerbl20233d}, and \cite{yu2024mip}, setting the learning rate for the normal vector at 0.3 for the Deep Blending dataset and 0.003 for the other datasets.
For the implementation of Scaffold-HGS, we did not increase the number of parameters in each Gaussian. Following the referenced paper, we employed an MLP to learn the normal vector based on the feature vector of each voxel and doubled the width of the output layer of the opacity MLP to accommodate two opacity values. The rasterizer was also modified based on Eq.~\ref{eq:10}. For HGS-MCMC, we adopt the hyperparameters and training settings from ~\cite{kheradmand20243d}, with the exception that we set the opacity threshold for Gaussian relocation to 0.015 for the Playroom scene. Additional training details are provided in the supplementary material.
All experiments were conducted using an NVIDIA RTX 3090 GPU.

\begin{figure}[H]
    \centering
\includegraphics[width=0.385\textwidth]{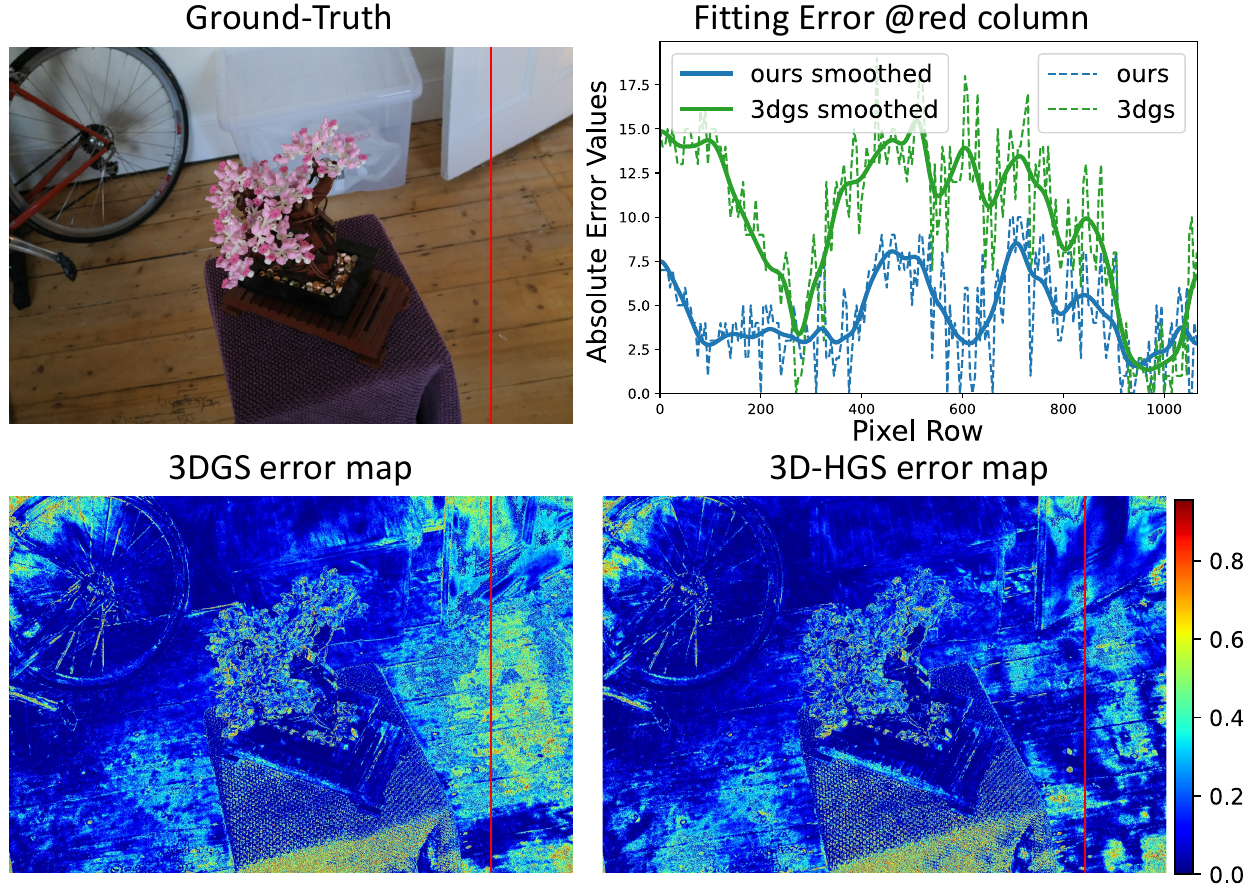} %
    \vspace{-0.3cm}
    \caption{{\bf Fitting Error.} Top-left: Frame from Bonsai scene. Top-right:  Fitting error using  3DGS and 3DHGS, along the red column. Bottom: Fitting error on the entire image, shown as a normalized heat map in the range [0,1] for 3DGS and 3DHGS.} 
    \vspace{-0.3cm}%
    \label{fig:bonsai} %
\end{figure}

\subsection{Results Analysis}\label{sec:exp}

\begin{figure*}[t]
\centering
\includegraphics[width=1\textwidth]{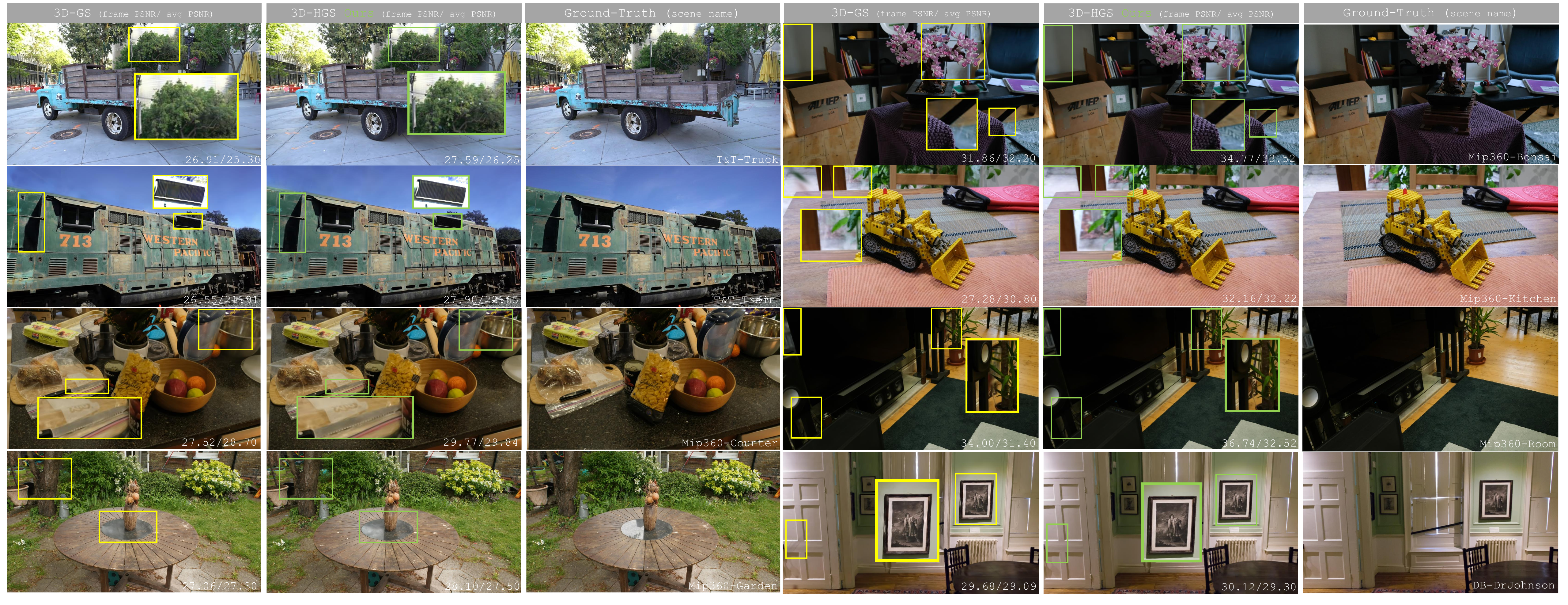}
\vspace{-0.4cm}
   \caption{\textbf{Qualitative comparison between 3D-GS and 3D-HGS.}
   \label{fig:visualresults}}
\vspace{-0.3cm}
\end{figure*}

\begin{figure}[h]
\centering
\includegraphics[width=0.48\textwidth]{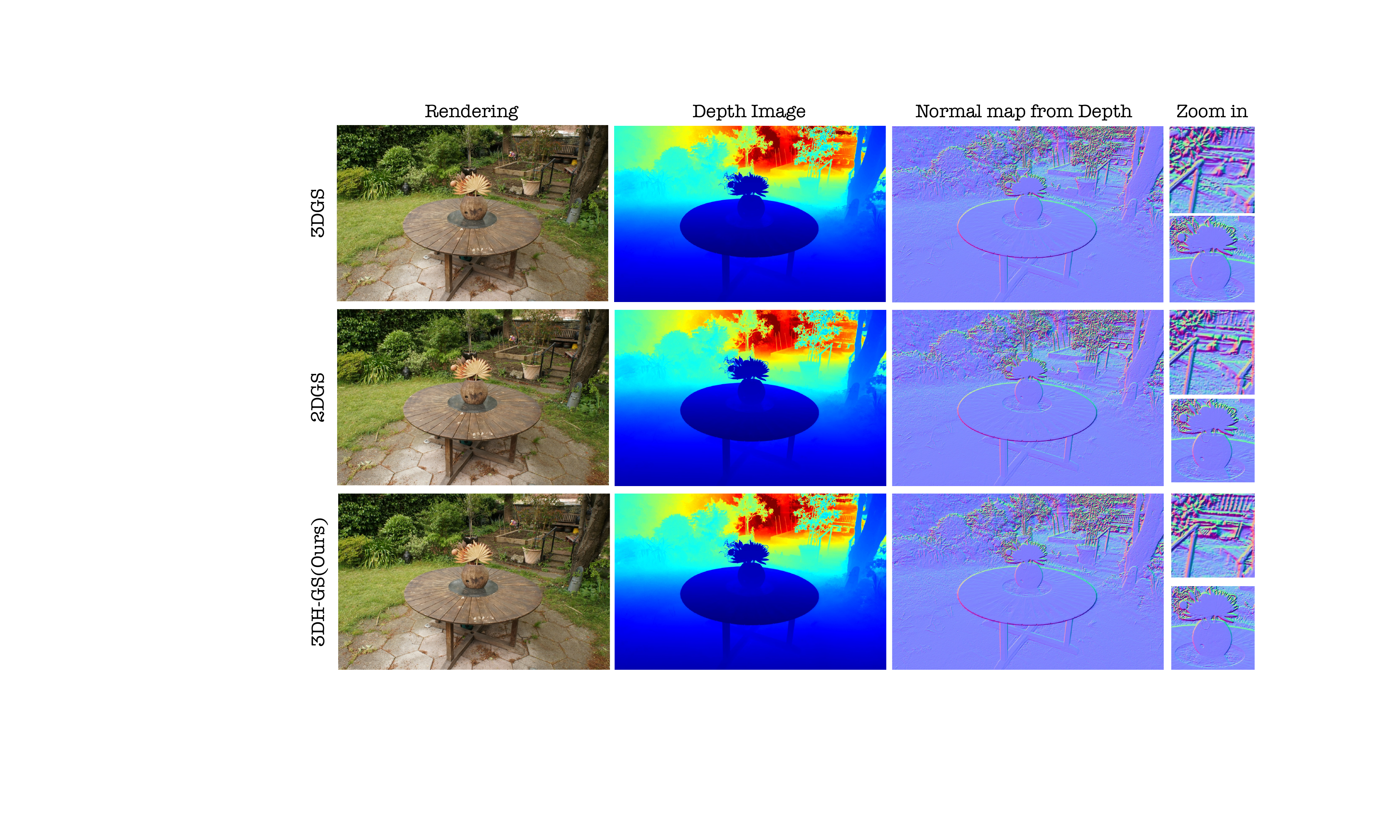}
\caption{\textbf{Comparison of Depth Images and Normal Maps.} We visualize the depth maps alongside normals estimated from these maps. Our method demonstrates superior performance, capturing finer details in the bench structure and the surface textures of pots.}\label{fig:normaleg}
\vspace{-0.5cm}
\end{figure}


\noindent{\bf Quantitative results}  are presented in Figs.~\ref{fig:comparisonHG} and \ref{fig:bonsai}, and in Tabs.~\ref{Tab:QuantaResults} and \ref{Tab:fpsmem}. Further details  are provided in the supplementary material. Across all datasets and methods, the integration of the 3D Half-Gaussian kernel yields substantial gains, establishing it as a superior choice for enhancing 3D novel view synthesis accuracy.

Fig.~\ref{fig:bonsai} shows plots of the grayscale fitting for the Bonsai scene along an image column containing both smooth surface regions and sharp transitions, along with the corresponding fitting error map for the entire image. The plots compare the proposed half-Gaussian kernel and standard 3D Gaussian Splatting (3DGS) against the ground truth.  Consistent with the results in Fig.~\ref{fig:square}, they illustrate that the half-Gaussian kernel achieves a more accurate fit than 3DGS, especially in areas with sharp transitions.

Table~\ref{Tab:QuantaResults} shows that incorporating the proposed Half Gaussian kernel, 3D-HGS, Scaffold-HGS, Mip-Splatting-HGS, and MCMC-HGS achieved SOTA performance across multiple datasets, consistently outperforming their respective baselines. This improvement underscores the effectiveness of the 3D Half-Gaussian kernel as a robust choice for the novel view synthesis. On the MipNeRF360 dataset, our proposed kernel enables 3D-HGS, Scaffold-HGS, Mip-Splatting-HGS, and MCMC-HGS to surpass baseline PSNRs by 0.78, 0.30, 0.49, and 0.24, respectively, with MCMC-HGS achieving a new state of the art. For the Tanks and Temples dataset, these methods see PSNR improvements of 0.85, 0.46, 0.78, and 0.77, again with MCMC-HGS setting a new benchmark. Lastly, on the Deep Blending dataset, our methods outperform baselines by 0.35, 0.15, 0.15, and 0.13 PSNR, with Scaffold-HGS achieving state-of-the-art performance. 

\begin{table}[H]
\caption{\textbf{Comparison of rendering speed and storage memory }between the proposed Half-Gaussian kernel-based methods and traditional Gaussian kernel-based methods.}
\begin{adjustbox}{width=\linewidth}
    \begin{tabular}{ l | c c | c c | c c }
    Dataset & \multicolumn{2}{c|}{Mip-NeRF360} & \multicolumn{2}{c|}{Tanks\&Temples} & \multicolumn{2}{c}{Deep Blending}  \\
    Method& FPS $\uparrow$ & Mem $\downarrow$ & FPS $\uparrow$ & Mem $\downarrow$  & FPS $\uparrow$ & Mem $\downarrow$    \\
    \cline{1-7}
    3D-GS \cite{kerbl20233d} &115 &762 & 149 &429 & 104& 668\\
    3D-\textbf{HGS} (\textbf{Ours})  &{ 125} &{ 694} & { 160} &437 & { 126} &{ 641} \\
    \cline{1-7}
    Scaffold-GS~\cite{lu2023scaffold} &120&173&120&77&129&55\\
    Scaffold-\textbf{HGS}(\textbf{Ours}) &118&180&115&84&{ 136}&{ 53}\\
    \cline{1-7}
    GS-MCMC~\cite{kheradmand20243d} &75& 732&133&438&90&969\\
    \textbf{HGS}-MCMC(\textbf{Ours}) &72&743&139&445&92&980 \\
    \cline{1-7}
    Mip-Splatting~\cite{kheradmand20243d} &76&970&117&569&91&843 \\
    Mip-Splatting-\textbf{HGS}(\textbf{Ours}) &83&883&121&566&102&808 \\

\end{tabular}
\end{adjustbox}
\label{Tab:fpsmem}
\end{table}

Finally, using half-Gaussian kernels does not significantly impact the rendering speed and memory requirements, as shown in Fig.~\ref{fig:comparisonHG}, Table~\ref{Tab:fpsmem}, {and in the ablation study in the supplemental material}.

\begin{table*}[t]
\caption{\textbf{PSNR scores $\uparrow$ of the rendering performance with different reconstruction kernels.} \textcolor{OrangeRed}{Red} indicates performance improvement, while \textcolor{cyan}{blue}  denotes performance decline. }
\vspace{-0.3cm}
\begin{adjustbox}{width=\linewidth}
    \begin{tabular}{ c |c c c| c c c| c c c c c c c c |c}
    \multirow{ 2}{*}{Kernels}& \multicolumn{3}{c|}{Tanks\&Temples} & \multicolumn{3}{c|}{Deep Blending} &\multicolumn{8}{c|}{Mip-NeRF360}&\multirow{ 2}{*}{AVG} \\
      & Truck & Train & Avg  & Playroom & Drjohnson & Avg  &  Bicycle&Garden&Kitchen&Stump& Room&Counter& Bonsai & Avg &   \\
    \hline
    3D-GS &25.30 &21.91 &23.60 &29.98&29.09  & 29.53&25.21&27.30&30.80&26.56&31.40&28.70&32.20 & {28.88} & {28.04}\\
    
    2D-GS& \textcolor{cyan}{25.14} & \textcolor{OrangeRed}{21.70}  & \cellcolor{orange!30}{23.42}& \textcolor{OrangeRed}{30.18}& \textcolor{OrangeRed}{29.12}& \cellcolor{yellow!30}{29.65} & \textcolor{cyan}{25.02} & \textcolor{cyan}{27.14} & \textcolor{OrangeRed}{31.33} & \textcolor{OrangeRed}{26.57} & \textcolor{OrangeRed}{31.37} & \textcolor{OrangeRed}{28.97} & \textcolor{OrangeRed}{32.33} & \cellcolor{orange!30}{28.94} & \cellcolor{orange!30}{28.06}\\

    GES & \textcolor{cyan}{24.94}  & \textcolor{OrangeRed}{21.73}&\cellcolor{yellow!30}{23.34} & \textcolor{OrangeRed}{30.29} & \textcolor{OrangeRed}{29.35} & \cellcolor{red!30}{29.82}& \textcolor{cyan}{24.87} &\textcolor{cyan}{27.07} &  \textcolor{OrangeRed}{31.07}&\textcolor{cyan}{26.17} & \textcolor{OrangeRed}{31.17}& \textcolor{OrangeRed}{28.75} & \textcolor{cyan}{31.97} & \cellcolor{yellow!30}28.72 & \cellcolor{yellow!30}27.94\\

    \hline
    
    3D-\textbf{HGS}&  \textcolor{OrangeRed}{26.25}& \textcolor{OrangeRed}{22.65}& \cellcolor{red!30}{24.45}& \textcolor{OrangeRed}{30.20}&\textcolor{OrangeRed}{29.30}& \cellcolor{orange!30}{29.76} &\textcolor{OrangeRed}{25.25}&\textcolor{OrangeRed}{27.50}&\textcolor{OrangeRed}{32.22}&\textcolor{OrangeRed}{26.64}&\textcolor{OrangeRed}{32.52}&\textcolor{OrangeRed}{29.84}&\textcolor{OrangeRed}{33.52}& \cellcolor{red!30}{29.66}&\cellcolor{red!30}{28.70}\\ 
    
 \end{tabular}
\end{adjustbox}
\vspace{-0.3cm}
\label{Tab:different_kernels_psnr}
\end{table*}


\noindent {\bf Qualitative results} are shown in  Figs.~\ref{fig:visualresults}, ~\ref{fig:normaleg}, {and in the supplemental material}. 
In Fig.~\ref{fig:visualresults} it can be  observed that our method delivers better performance on fine-scale details (e.g., T\&T-Truck, Mip360-Garden, Mip360-Bonsai, Mip360-Room), high-frequency textures (e.g., T\&T-Train, Mip360-Counter), light rendering (e.g., Mip360-Garden, DB-DrJohnson), and shadow areas (e.g., T\&T-Train, Mip360-Bonsai, DB-DrJohnson).
Fig.~\ref{fig:normaleg} provides an example of a generated depth image and the corresponding estimated surface normals, which were produced using the method  described in \cite{huang20242d}. The proposed method captures better finer details in the bench structure and the surface texture of the pots.

\noindent {\bf Different Reconstruction Kernels} Table~\ref{Tab:different_kernels_psnr} provides
the PSNR score for each 3D scene, the average score for each dataset, and the total average score when using the different kernels.  In this experiment, we use 3D-GS as the baseline, with improved and degraded results highlighted in red and  blue, respectively. Compared to the original 3D Gaussian kernel, the 3D Half-Gaussian kernel shows consistent improvement across all 3D scenes. While other kernels demonstrate superiority in some 3D scenes, they exhibit drawbacks in others. Overall, we achieved the best average performance across all 3D scenes.

%% file: sec/5conclusion.tex
\section{Conclusion}

We introduced Half-Gaussian Splatting, a  plug-and-play method for accurate 3D  view synthesis. Our approach uses two distinct opacities in each Gaussian, allowing precise rendering control without increasing inference time. This design achieves SOTA performance across multiple datasets and integrates seamlessly into most Gaussian-splatting architectures without structural changes. We validated the effectiveness of our approach across four baseline methods, consistently improving accuracy. To rigorously assess its advantages, we compared it against other kernel modification techniques, confirming Half-Gaussian Splatting as a highly effective choice for 3D splatting-based methods. 

%% file: sec/X_suppl.tex
\clearpage
\setcounter{page}{1}
\maketitlesupplementary

\renewcommand{\thesection}{\Alph{section}}
\setcounter{section}{0} 

\section{Detailed Derivations}
\subsection{3D Half-Gaussian Kernel}\label{}
In this section, we give a more detailed derivation for Eq.~\ref{eq:7}, and Eq.~\ref{eq:8}. 
We start with the density function of a 3D Half-Gaussian with covariance matrix \( \Sigma \):
\begin{equation}
HG_\mathbf{\Sigma}(\textbf{x}) = 
    \left\{
    \begin{array}{ll}
    e^{-\frac{1}{2}\textbf{x}^T\boldsymbol{\Sigma}^{-1}\textbf{x}} &  \mathbf{n}^T\mathbf{x} \ge 0\\
    0 & \mathbf{n}^T\mathbf{x} <0\\
    \end{array}
    \right. 
\end{equation}
 To compute the integral of the non-zero Half Gaussian along the $z$ direction, $ \int_{\mathbf{n}^T\mathbf{x} \ge 0}{HG}_{\mathbf{\Sigma}}(\mathbf{x}) dz$, 
we start by calculating the conditional distribution of $z$, given $x$ and $y$ and the marginal distribution of $x$ and $y$:

\begin{equation}
    f(z|x,y)=\exp\left(-\frac{1}{2}\left(\frac{z-\mu_{z|xy}}{\sigma_{zz|xy}}\right)^2\right)
\end{equation}

\begin{equation}
f(x,y) = \exp\left(-\frac{1}{2}\begin{bmatrix}x \ y \end{bmatrix} \Sigma_{xx} ^{-1} \begin{bmatrix}x \\ y \end{bmatrix}\right)
\end{equation}
where $\mu_{z|xy}$ and $\sigma_{zz|xy}$ are the mean and variance of the conditional distribution, respectively. $\Sigma_{xx}$ is defined as the top left $2 \times 2$ covariance submatrix of $\Sigma$:

\begin{equation}
\Sigma = \begin{bmatrix} \Sigma_{xx} & \Sigma_{xz} \\ \Sigma_{zx} & \sigma_{zz} \end{bmatrix}
\end{equation}
and \( \sigma_{zz} \) is the variance for \( z \). Then $\mu_{z|xy}$ and $\sigma_{zz|xy}$ can be expressed as:
\begin{equation}
    \mu_{z|xy} =  \Sigma_{zx} \Sigma_{xx}^{-1} \begin{bmatrix}
        x \\ y
    \end{bmatrix}
\end{equation}

\begin{equation}
    \sigma_{zz|xy} = \sigma_{zz} - \Sigma_{zx} \Sigma_{xx}^{-1} \Sigma_{xz}
\end{equation}
Then, the integral of the 3D Half-Gaussian can be computed as: 
\begin{multline}
    \int_{\mathbf{n}^T\mathbf{x} \ge 0}{HG}_{\mathbf{\Sigma}}(\mathbf{x}) dz =  f(x,y) \int^{\infty}_{-\frac{n_1x+n_2y}{n_3}}f(z|x,y)dz  \\
    =  \exp\left(-\frac{1}{2}\begin{bmatrix}x \ y \end{bmatrix} \Sigma_{xx} ^{-1} \begin{bmatrix}x \\ y \end{bmatrix}\right)
    \\  \int^{\infty}_{-\frac{n_1x+n_2y}{n_3}} \exp\left(-\frac{1}{2}\left(\frac{z-\mu_{z|xy}}{\sigma_{zz|xy}}\right)^2\right) dz
\end{multline}
where $\exp\left(-\frac{1}{2}\begin{bmatrix}x \ y \end{bmatrix} \Sigma_{xx} ^{-1} \begin{bmatrix}x \\ y \end{bmatrix}\right)$ is the same as $\hat{G}_{\mathbf{\hat{\Sigma}}}(\hat{\mathbf{x}}-\mathbf{\hat{\mu}})$. We apply the substitution $u=\frac{z-\mu_{z|xy}}{\sqrt{2}\sigma_{zz|xy}}$, then we get:

\begin{multline}
    \int_{\mathbf{n}^T\mathbf{x} \ge 0}{HG}_{\mathbf{\Sigma}}(\mathbf{x}) dz  \\
    =  \hat{G}_{\mathbf{\hat{\Sigma}}}(\hat{\mathbf{x}}-\mathbf{\hat{\mu}})
    \int^{\infty}_{\frac{-\frac{n_1x+n_2y}{n3}-\mu_{z|xy}}{\sqrt{2}\sigma_{zz|xy}}}e^{-u^2}du 
\end{multline}

The final result of the integral of the 3D Half-Gaussian is:
\begin{multline}
  \int_{\mathbf{n}^T\mathbf{x} \ge 0}{HG}_{\mathbf{\Sigma}}(\mathbf{x}) dz \\  =  \hat{G}_{\mathbf{\hat{\Sigma}}}(\hat{\mathbf{x}}-\mathbf{\hat{\mu}})
\frac{1}{2} \operatorname{erfc} \left( 
-\frac{
\left(\begin{bmatrix} n_1 \  n_2\end{bmatrix}/n_3+ \Sigma_{zx} \Sigma_{xx}^{-1}\right) \begin{bmatrix} x \\ y \end{bmatrix}}{\sqrt{2 \left( \sigma_{zz} - \Sigma_{zx} \Sigma_{xx}^{-1} \Sigma_{xz} \right)}} 
\right)
\end{multline}

\subsection{Fourier transform of 1D Half-Gaussian}\label{}

\begin{table*}[h]
\caption{\textbf{{Quantitative comparison to previous methods on standard benchmarks. Competing metrics are directly taken from the respective papers for consistency.}}}
\begin{adjustbox}{width=\linewidth}
    \begin{tabular}{ ccccccc }
     \hline
     & & 3DGS & 3DHGS & Scaffold-HGS & 3DHGS-Mip-Splatting & 3DHGS-MCMC \\ 
    & &\tiny PSNR$\uparrow$/SSIM$\uparrow$/LPIPS$\downarrow$ &\tiny PSNR$\uparrow$/SSIM$\uparrow$/LPIPS$\downarrow$ &\tiny PSNR$\uparrow$/SSIM$\uparrow$/LPIPS$\downarrow$ &\tiny PSNR$\uparrow$/SSIM$\uparrow$/LPIPS$\downarrow$&\tiny PSNR$\uparrow$/SSIM$\uparrow$/LPIPS$\downarrow$ \\
    \hline
    \multirow{8}{*}{\rotatebox[origin=c]{90}{ MipNeRF360}}&
     Counter &28.70/0.905/0.204  &29.84/0.920/0.177 & 29.52/0.914/0.183&30.11/0.925/0.166 & 30.24/0.928/0.161 \\
    & Stump &26.56/0.770/0.217 &26.64/0.760/0.242 &26.45/0.76/0.259 & 26.94/0.777/0.215& 27.17/0.794/0.202\\
    & Kitchen &30.80/0.927/0.127 &32.22/0.936/0.113 &31.67/0.931/0.113 &32.56/0.938/0.108 & 32.72/0.940/0.108\\
    & Bicycle &25.21/0.765/0.209 &25.25/0.750/0.230 & 25.05/0.74/0.258&25.43/0.773/0.189 & 25.52/0.777/0.196\\
    & Bonsai &32.20/0.946/0.183 &33.52/0.950/0.180 & 32.75/0.947/0.176&33.55/0.952/0.161 & 34.33/0.958/0.159\\
    & Room &31.40/0.918/0.223 &32.52/0.931/0.193 & 32.21/0.929/0.184&32.80/0.937/0.175 & 33.25/0.940/0.169\\
    & Garden &27.30/0.865/0.107 &27.50/0.860/0.110 & 27.12/0.848/0.130&27.80/0.865/0.106 & 27.67/0.864/0.110\\
    \cline{2-7}
    & Average &28.88/0.870/0.182 &29.66/0.873/0.178 & 29.25/0.867/0.186 &29.88/0.881/0.160 &30.13/0.886/0.158\\
    \cline{2-7}
    &Flowers &21.52/0.605/0.336 &21.55/0.610/0.250 & /& /& /\\
    &Treehill &22.49/0.638/0.317 &22.63/0.640/0.330 & /& /& /\\
    \hline
    \multirow{3}{*}{\rotatebox[origin=c]{90}{ \scriptsize \makecell{ Tanks \\ \& Temples}}}&Train &21.91/0.815/0.210 &22.65/0.826/0.193 & 22.63/0.828/0.186&22.65/0.836/0.176 &23.24/0.841/0.180\\
    & Truck&25.30/0.88/0.152 &26.25/0.887/0.138 &26.17/0.889/0.138 &26.42/0.894/0.115 &26.91/0.902/0.107 \\
    \cline{2-7}
    & Average&23.60/0.847/0.181 &24.45/0.857/0.169 & 24.42/0.859/0.162&24.53/0.865/0.145 &25.08/0.841/0.144\\
    \hline 

    \multirow{3}{*}{\rotatebox[origin=c]{90}{ \scriptsize \makecell{ Deep \\ Blending}}}&Dr Johnson &29.02/0.902/0.244 &29.30/0.903/0.240 & 29.82/0.908/0.236& 29.18/0.901/0.241& 29.02/0.890/0.26 \\
    & Playroom&29.81/0.904/0.242 & 30.22/0.907/0.244&30.90/0.912/0.244 &30.04/0.901/0.241 & 30.58/0.905/0.231 \\
    \cline{2-7}
    & Average&29.41/0.903/0.243& 29.76/0.905/0.242&30.36/0.910/0.240 &29.61/0.901/0.241 & 29.80/0.898/0.245\\
    \hline 
    \end{tabular}
\end{adjustbox}
\label{tab:all}
\end{table*}

In this section, we present a detailed derivation of the Fourier transform for the 1D Half-Gaussian, which was utilized to generate Fig.~\ref{fig:sqright}. We begin with the definition of the half-Gaussian:

\begin{equation}
f(x) = 
\begin{cases} 
e^{-\frac{x^2}{2\sigma^2}}, & x \geq 0 \\ 
0, & x < 0 
\end{cases}
\end{equation}

The following provides the derivation of its Fourier transform:

\begin{align}
    g(k) & = \int_{-\infty}^\infty HG(x) e^{-ikx} dx \\
      & = \int_{0}^\infty e^{-\frac{x^2}{2\sigma^2}} e^{-ikx} dx \\
      & = \int_{0}^\infty e^{\frac{-1}{2\sigma^2} (x^2 + 2i\sigma^2kx)} dx \\
      & = \int_{0}^\infty e^{\frac{-1}{2\sigma^2} ((x + i\sigma^2k)^2 - (i\sigma^2k)^2)} dx \\ 
      &  = e^{-\frac{\sigma^2k^2}{2}} \int_{0}^\infty e^{-\frac{1}{2\sigma^2} (x + i\sigma^2k)^2} dx
\end{align}

Now we apply the substitution \( u = x + i\sigma^2k \), where \( du = dx \). Under this substitution, the integration limits transform. The integral becomes:

\begin{equation}
g(k) = e^{-\frac{\sigma^2 k^2}{2}} \int_{i\sigma^2k}^\infty e^{-\frac{u^2}{2\sigma^2}} du
\end{equation}

The remaining integral can be evaluated using the complex error function \( \text{erfi}(z) \), resulting in the final expression:

\begin{equation}
g(k) = e^{-\frac{\sigma^2 k^2}{2}} \sqrt{\frac{\pi}{2}} \left( \sigma - i\sigma \, \text{erfi}\left(\frac{k\sigma}{\sqrt{2}}\right) \right)
\end{equation}

This provides the Fourier transform of the half-Gaussian function.

\section{Implementation Details}\label{}
\subsection{Splatting for Splitting Plane}

For simplicity, we define the splitting plane as intersecting the 3D ellipsoid at the three-sigma boundary of the original Gaussian and oriented orthogonal to the normal direction. To approximate the splitting plane, we use an external ellipsoid that circumscribes the original 3D ellipsoid, with a significantly shortened third axis aligned parallel to the normal of the splitting plane.

During rendering, the precomputed 2D projection of the splitting plane facilitates efficient determination of the inner boundary of the Gaussian. Combined with the outer boundary derived from the original Gaussian, this enables accurate partitioning into two halves and precise identification of the valid effect areas for each half-Gaussian. This approach achieves rendering speeds that are comparable to or even exceed those of standard Gaussian splatting.

\begin{algorithm}
\caption{Covariance Matrix Transformation \newline $\mathbf{\Sigma}$: The covariance matrix for the 3D gaussian \newline $\mathbf{n}$: The normal for the cutting plane \newline $\mathbf{\Sigma}$: the covariance for the ellipsoid include the cutting plane
}
\begin{algorithmic}
  
        \STATE Normalize the normal vector: $\mathbf{n} \leftarrow \mathbf{n} / \|\mathbf{n}\|$
    
        \IF{$n_x = 0$ and $n_y = 0$}
            \STATE Set $\mathbf{v}_1 = (1, 0, 0)$ and $\mathbf{v}_2 = (0, 1, 0)$
        \ELSE
            \STATE $\mathbf{v}_1 \leftarrow (n_y, -n_x, 0) / \|(n_y, -n_x, 0)\|$
            \STATE $\mathbf{v}_2 \leftarrow \mathbf{n} \times \mathbf{v}_1 / \|\mathbf{n} \times \mathbf{v}_1\|$
        \ENDIF
    
        \STATE $\text{basis} \leftarrow \text{concatenate}(\mathbf{v}_1, \mathbf{v}_2, \mathbf{n})$
    
        \STATE $\text{cov} \leftarrow \text{basis}^T \cdot \Sigma \cdot \text{basis}$
        
        \STATE $a' \leftarrow \text{cov}[0,0] - \frac{\text{cov}[0,2]^2}{\text{cov}[2,2]}$
        
        \STATE $b' \leftarrow \text{cov}[0,1] - \frac{\text{cov}[0,2]\text{cov}[1,2]}{\text{cov}[2,2]}$
        
        \STATE $c' \leftarrow \text{cov}[1,1] - \frac{\text{cov}[1,2]\text{cov}[1,2]}{\text{cov}[2,2]}$
    
        \STATE $\sigma \leftarrow \begin{bmatrix} a' & b' & 0 \\ b' & c' & 0 \\ 0 & 0 & \min(a', c') / 100 \end{bmatrix}$
    
        \STATE $\sigma \leftarrow \text{basis} \cdot \sigma \cdot \text{basis}^T$
\end{algorithmic}
\label{alg:small_gs}
\end{algorithm}

Additionally, we provide a detailed procedure to compute the covariance matrix for the smaller circumscribing ellipsoid based on the covariance matrix of the original Gaussian and the normal vector of the cutting plane in Algorithm~\ref{alg:small_gs}.

\subsection{Training Settings}
During training, we introduce a new parameter, the normal vector, and assign it a dedicated learning rate of 0.003. Unlike original Gaussian Splatting, each Gaussian in our framework has two opacities. The initial learning rates for opacities are consistent with those used in Gaussian Splatting. In 3D-HGS, the learning rates for both opacity and the normal vector are progressively reduced by a factor of 1.4 every 5000 iterations.

To further improve training efficiency, we adjust the thresholds for opacity during the densification and pruning stages. Specifically, we increase the opacity threshold to 0.01 and reset the opacity value for half-Gaussians to 0.02. These adjustments help to effectively eliminate dust and noise from the scene, leading to cleaner and more accurate representations.
\section{Additional Experiments}
\subsection{Detailed Results}
We evaluated the performance of 3D-HGS and other 3D Gaussian architectures, including Scaffold, mip-splatting, and MCMC, using our 3D half-Gaussian kernel and compared to the original 3D Gaussian Splatting. Table~\ref{Tab:QuantaResults} {in the main paper} summarizes the average results derived from Table~\ref{tab:all}. Our method consistently outperforms the original 3D Gaussian Splatting across all metrics. Each average reported in Table~\ref{tab:all} was computed across all scenes within the dataset. 

\subsection{Splitting Plane Visualization}
\begin{figure}[h]
    \centering
\includegraphics[width=0.3\textwidth]{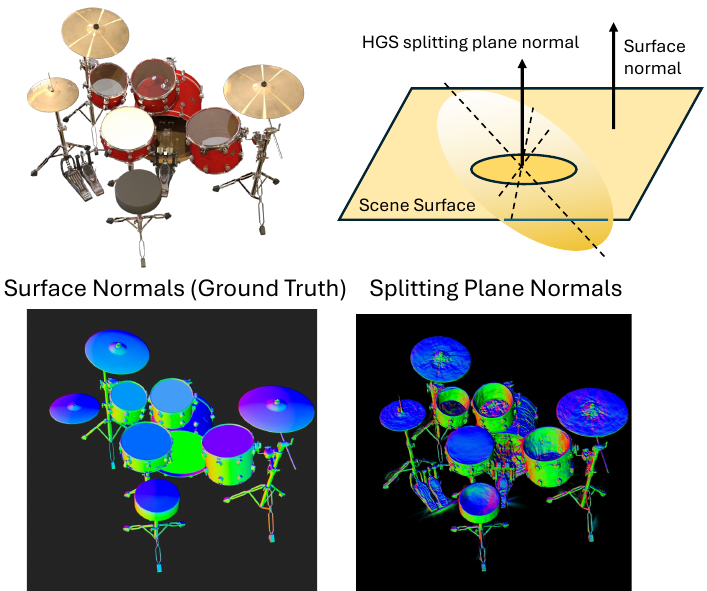} %
    \vspace{-0.25cm}
    \caption{{\bf Splitting Plane Visualization.}  The normals to the  splitting planes are parallel to the corresponding 3D surface normals. Please note that the tops of three of the drums are transparent.}   %
    \label{fig:normal_map} %
    \vspace{-0.25cm}
\end{figure}
{In order to visualize the locations of the splitting planes, we provide side by side the scene surface normals and the splitting plane normals for the drums scene in Fig.~\ref{fig:normal_map}}. The figure shows that the normals of the splitting planes generally align well with the  normals of the scene surfaces, indicating that the half ellipsoids  effectively model the sharp transitions between the object surfaces  and the empty space. This is also supported by Fig.~\ref{fig:hist} in the main paper, which shows that a large number of Gaussians have one of their halves transparent.

\subsection{Training time vs PSNR}
\begin{figure}[h!]
\vspace{-0.2cm}
    \centering
    \includegraphics[width=0.4\textwidth]{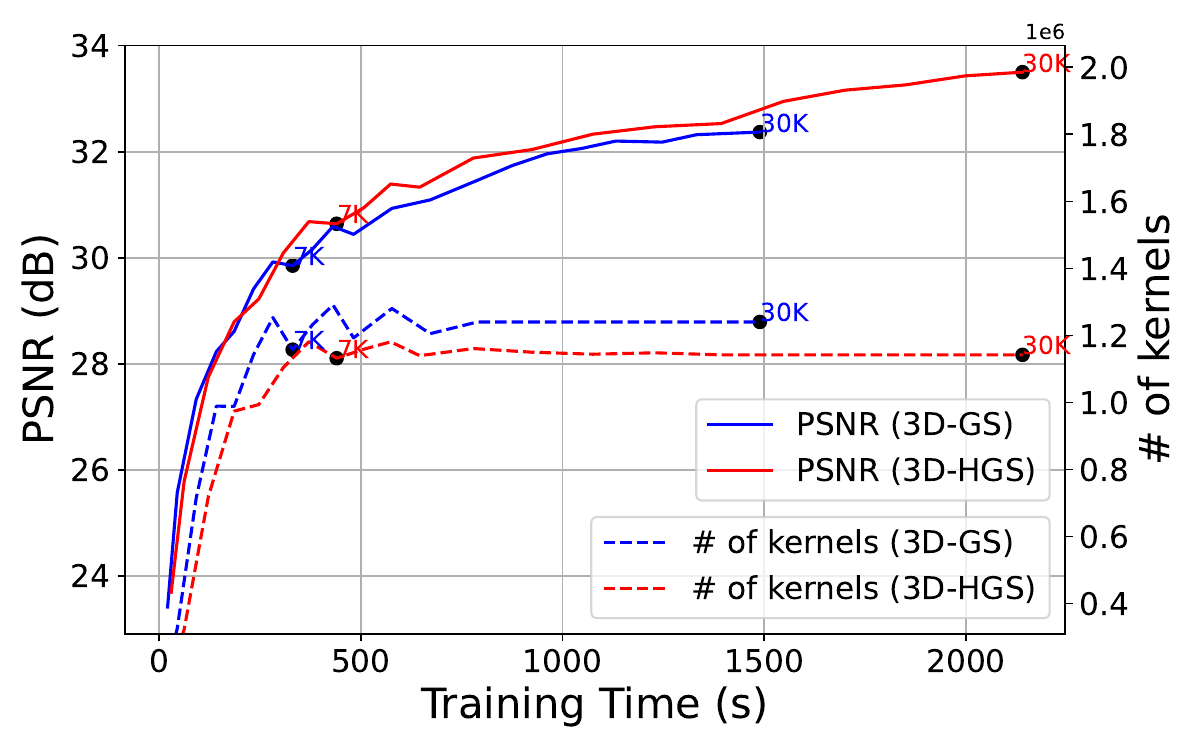} %
    \vspace{-0.4cm}
    \caption{{\bf Training Time:} Performance (PSNR$\uparrow$) and memory (number of kernels$\downarrow$) versus training time for the Bonsai scene.  Numbers on the curves indicate number of iterations. }   %
    \label{fig:traintime} %
\end{figure}
As shown in  Fig.~\ref{fig:traintime}, for the  same training time,  3DHGS using a similar or smaller number of Gaussians achieves better PSNR than using 3DGS. 

\subsection{Rendering Speed Ablation Study}
\begin{table}[h!]
\vspace{-0.2cm}
\caption{{\bf Ablation study} on Rendering Speed  on a single RTX-3090. A 3DGHS kernel consists of its two half-Gaussians.}
\label{tab:trainspeed}
\vspace{-0.2cm}
\centering
\begin{adjustbox}{width=\linewidth}
\begin{tabular}{@{}lcc@{}ccc@{}}
\toprule
Dataset            & \multicolumn{2}{c}{\small \# of Kernels}  & & \multicolumn{2}{c}{\small 3DHGS FPS / PSNR}  \\ 
\cmidrule(lr){2-3}   \cmidrule(lr){5-6} 
& 3DGS      & 3DHGS   
&  &w/o Eff. Rasterizer & with Eff. Rasterizer \\
\midrule
\small Mip-NeRF 360       & 3.22 M  & 2.88 M & & 76 / 29.56 & 125 / 29.66\\
\small Tank \& Template   & 1.81 M & 1.81 M & & 95 / 24.49 & 160 / 24.45 \\
\small Deep Blending      & 3.00 M & 2.66 M & & 82 / 29.76  & 126 / 29.76        \\ \bottomrule
\end{tabular}
\end{adjustbox}
\vspace{-0.25cm}
\end{table}

We performed an ablation study on the proposed efficient 3D half-Gaussian Splatting Rasterizer (Sec.\ref{3.4}). Table~\ref{tab:trainspeed} reports the rendering speed in FPS and the performance in PSNR for our method, 3DHGS, with and without the efficient Rasterizer. The results highlight a significant improvement in rendering speed with our proposed Rasterizer. Additionally, we provide the average number of kernels used by 3DGS and 3DHGS for reference.